\documentclass[10pt,twocolumn,letterpaper]{article}

\usepackage[pagenumbers]{cvpr}

\usepackage[dvipsnames]{xcolor}

\definecolor{cvprblue}{rgb}{0.21,0.49,0.74}
\usepackage[pagebackref,break links,color links,citecolor=cvprblue]{hyperref}
\usepackage{bm}
\usepackage{color}
\usepackage{floatrow}
\newfloatcommand{capbtabbox}{table}[][\FBwidth]

\newcommand\nonumfootnote[1]{
\begingroup
    \renewcommand\thefootnote{}\footnote{\hspace{-3.5pt}#1}
    \addtocounter{footnote}{-1}
\endgroup
}

\newcommand{\method}{\textcolor{black}{\mbox{\texttt{SparseCtrl}}}\xspace}
\def\para#1{\vspace{0.25em}\noindent\textbf{#1}}

\title{
SparseCtrl: Adding Sparse Controls to Text-to-Video Diffusion Models
\vspace{-10pt}
}

\author{
    Yuwei Guo$^{1}$\:
    Ceyuan Yang$^{2\dag}$\:
    Anyi Rao$^3$\:
    Maneesh Agrawala$^3$\:
    Dahua Lin$^{1,2}$\:
    Bo Dai$^2$ \\
    \small{
    $^1$The Chinese University of Hong Kong\quad
    $^2$Shanghai Artificial Intelligence Laboratory\quad
    $^3$Stanford University
    } \\
    \small{\texttt{\{gy023,dhlin\}@ie.cuhk.edu.hk}} \quad
    \small{\texttt{\{yangceyuan,daibo\}@pjlab.org.cn}} \vspace{-3pt} \\
    \small{\texttt{\{anyirao,maneesh\}@cs.stanford.edu}}
}

\begin{document}

\twocolumn[{
\renewcommand\twocolumn[1][]{#1}
\maketitle
\begin{center}
    \centering
    \captionsetup{type=figure}
    \vspace{-20pt}
    \includegraphics[width=0.9\textwidth]{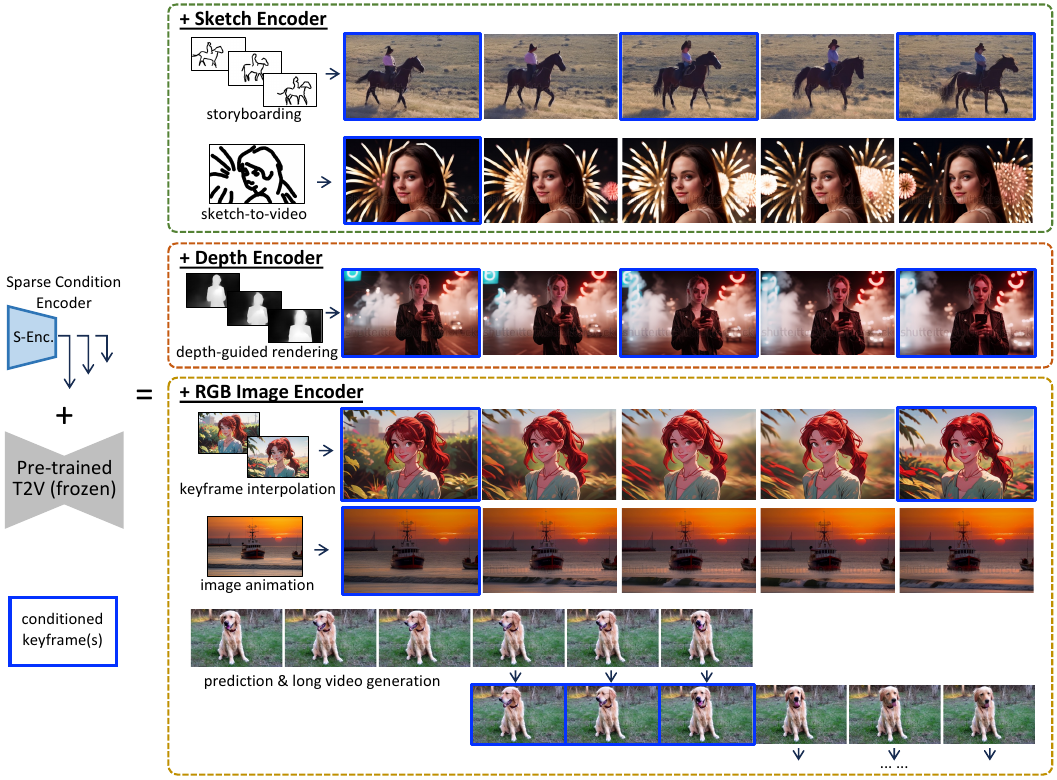}
    \vspace{-5pt}
    \captionof{figure}{
        We present \method, an add-on encoder network upon pre-trained text-to-video (T2V) diffusion models to accept additional temporally sparse conditions for specific keyframe(s), \emph{e.g.}, sketch/depth/RGB image.
        Through integration with various modality encoders, \method enables the pre-trained T2V for various applications including storyboarding, sketch-to-video, image animation, long video generation, etc.
        When combined with AnimateDiff~\cite{guo2023animatediff} and enhanced personalized image backbones~\cite{realisticvision, toonyou}, \method also achieves controllable, high-quality generation results, as shown in the 2/3/4-th rows.
    }
    \label{fig:teaser}
\end{center}
}]

\nonumfootnote{$^\dag$Corresponding Author.}

\begin{abstract}
The development of text-to-video (T2V), \emph{i.e.}, generating videos with a given text prompt, has been significantly advanced in recent years.
However, relying solely on text prompts often results in ambiguous frame composition due to spatial uncertainty.
The research community thus leverages the \texttt{dense} structure signals, \emph{e.g.}, per-frame depth/edge sequences to enhance controllability, whose collection accordingly increases the burden of inference.
In this work, we present \method to enable flexible structure control with temporally \texttt{sparse} signals, requiring only one or few inputs, as shown in~\cref{fig:teaser}.
It incorporates an additional condition encoder to process these sparse signals while leaving the pre-trained T2V model \texttt{untouched}.
The proposed approach is compatible with various modalities, including sketches, depth, and RGB images, providing more practical control for video generation and promoting applications such as storyboarding, depth rendering, keyframe animation, and interpolation.
Extensive experiments demonstrate the generalization of \method on both original and personalized T2V generators.
Codes and models will be publicly available at
\urlstyle{same}\url{https://guoyww.github.io/projects/SparseCtrl}.

\vspace{-10pt}
\end{abstract}
\section{Introduction}
With the advance of text-to-image (T2I) generation~\cite{saharia2022photorealistic, balaji2022ediffi, gu2022vector, kumari2023multi, xu2023versatile, wu2023harnessing, podell2023sdxl, ramesh2022hierarchical, dai2023emu} and large-scale text-video paired datasets~\cite{bain2021frozen},  there has been a surge of progress in the field of text-to-video (T2V) generative models~\cite{singer2022make, ho2022imagen, blattmann2023align}.
These developments enable users to generate compelling videos through textual descriptions of the desired content.
Nonetheless, textual prompts, being inherently abstract expressions, struggle to accurately define complex structural attributes such as spatial layouts, poses, and shapes.
This lack of precise control impedes its practical application in more demanding and professional contexts, such as anime creation and filmmaking.
Consequently, users often find themselves engrossed in numerous rounds of random trial-and-error to achieve their desired outputs. 
This process can be time-consuming, especially since there is no straightforward method to guide the synthetic results toward the expected direction during the iterative trying process.
To unlock the potential of T2V generation, efforts have been made to incorporate more precise control through structural information.
For instance, Gen-1~\cite{esser2023structure} pioneers using monocular depth maps as structural guidance.
VideoComposer~\cite{wang2023videocomposer} and DragNUWA~\cite{yin2023dragnuwa} investigate the domain of compositional video generation, employing diverse modalities such as depth, sketch, and initial image as control signals.
Furthermore, prior studies~\cite{khachatryan2023text2video, zhang2023controlvideo, guo2023animatediff} utilize the image ControlNet~\cite{zhang2023adding} to introduce various controlling modalities to video generation.
By harnessing additional structural sequences, these approaches provide enhanced control capabilities.
However, for precise output control, existing works necessitate temporally \emph{dense} structural map sequences, which means that users need to furnish condition maps for each frame in the generated video, thereby increasing the practical costs.
Additionally, most approaches towards controllable T2V typically redesign the model architecture to accommodate the extra condition input, which demands costly model retraining.
Such practice is inefficient when a well-trained T2V model is already available or when there is a requirement to incorporate a new control modality into a pre-trained generator.

In this paper, we introduce \method, an efficient approach that targets controlling text-to-video generation via temporally sparse condition maps with an add-on encoder.
More specifically, in order to control the synthesis, we apply the philosophy of ControlNet~\cite{zhang2023adding}, which implements an auxiliary encoder while preserving the integrity of the original generator.
This design allows us to incorporate additional conditions by merely training the encoder network on top of the pre-trained T2V model, thereby eliminating the need for comprehensive model retraining.
Additionally, this design facilitates control over not only the original T2V but also the derived personalized models when combined with the plug-and-play motion module of AnimateDiff~\cite{guo2023animatediff}.
To achieve this, we design a condition encoder equipped with temporal-aware layers that propagate the sparse condition signals from conditioned keyframes to unconditioned frames.
Significantly, we find that purging the noised sample input in the vanilla ControlNet further prevents potential quality degradation in our scenario.
Moreover, we apply widely used masking strategies~\cite{yu2023magvit, blattmann2023align, zhang2023show, chen2023seine} during training to accommodate varying degrees of sparsity and tackle a broad range of application scenarios.

We evaluate \method by training three encoders on sketches, depth, and RGB images. 
Experimental results show that users can manipulate the structure of the synthetic videos by providing just one or a few input condition maps.
Comprehensive ablation studies are performed to investigate the contribution of each component.
We additionally show that by integrating with plug-and-play video generation backbone such as AniamteDiff~\cite{guo2023animatediff}, our method exhibits compatibility and excellent visual quality with various personalized text-to-image models.
Leveraging this sparse control approach, \method enables a broad range of applications.
For instance, the sketch encoder empowers users to transform hand-drawn storyboards into dynamic videos;
The depth encoder provides the ability to render videos by supplying a minimum number of depth maps;
Furthermore, the RGB image encoder unifies multiple tasks, including image animation, keyframe interpolation, video prediction, etc.
We anticipate that this work will contribute towards bridging the gap between text-to-video research and real-world content creation processes.

\section{Related Works}

\para{Text-to-video diffusion models.}
The field of text-to-video (T2V) generation~\cite{karras2023dreampose, ruan2023mm, zhang2023i2vgen, he2022latent, chen2023videocrafter1, hong2022cogvideo, girdhar2023emu, wang2023modelscope, an2023latent, gu2023reuse} has witnessed significant progression recently, driven by advancements in diffusion models~\cite{sohl2015deep, ho2020denoising, song2020denoising, dhariwal2021diffusion} and large-scale text-video paired datasets~\cite{bain2021frozen}.
Initial attempts in this area focus on training a T2V model from scratch.
For example, Video Diffusion Model~\cite{ho2022video} expands the standard image architecture to accommodate video data and trains on both image and video together.
Imagen Video~\cite{ho2022imagen} employs a cascading structure for high-resolution T2V generation, while Make-A-Video~\cite{singer2022make} uses a text-image prior model to reduce reliance on text-video paired data.
Others turn to build T2V models upon powerful text-to-image (T2I) models such as Stable Diffusion~\cite{podell2023sdxl}, by incorporating additional layers to model cross-frame motion and consistency~\cite{zhou2022magicvideo, ge2023preserve, wang2023lavie}.
Among these, MagicVideo~\cite{zhou2022magicvideo} utilizes a causal design and executes training in a compressed latent space to mitigate computational demands.
Align-Your-Latents~\cite{blattmann2023align} efficiently turns T2I into video generators by aligning independently sampled noise maps.
AnimateDiff~\cite{guo2023animatediff} utilizes a pluggable motion module to enable high-quality animation creation on personalized image backbones~\cite{ruiz2023dreambooth, gal2022image, kumari2023multi, ruiz2023hyperdreambooth}.
Other contributions include noise prior modeling~\cite{ge2023preserve}, training on high-quality datasets~\cite{wang2023lavie}, and latent-pixel hybrid space denoising~\cite{zhang2023show}, all leading to remarkable pixel quality.
However, current text-conditioned video generation techniques lack fine-grained controllability over synthetic results.
In response to this challenge, our work aims to enhance the control of T2V models via an add-on encoder.

\para{Controllable text-to-video generation.}
Given that a text prompt can often result in ambiguous guidance to the video motion, content, and spatial structure, such controllabilities become crucial factors in T2V generation.
For high-level video motion control, several studies propose learning LoRA~\cite{hu2021lora} layers for specific motion patterns~\cite{guo2023animatediff, zhao2023motiondirector}, while others employ extracted trajectories~\cite{yin2023dragnuwa}, motion vectors~\cite{wang2023videocomposer}, or pose sequence~\cite{ma2023follow}.
To manage specific synthetic keyframes for animation or interpolation, recent explorations include encoding the image separately to the generator~\cite{xing2023dynamicrafter}, concatenating with the noise input~\cite{chen2023seine, wang2023videocomposer}, or utilizing multi-level feature injection~\cite{yin2023dragnuwa}.
For fine-grained spatial structure control, some low-level representations are introduced.
Gen-1~\cite{esser2023structure} is the first to use monocular depth sequences as structural guidance.
VideoComposer~\cite{wang2023videocomposer} encodes sketch and depth sequences via a shared encoder, facilitating flexible combinations at inference.
Additionally, some approaches utilize readily available image controlling models~\cite{zhang2023adding, mou2023t2i} for controllable video generation~\cite{chen2023control, zhang2023controlvideo, khachatryan2023text2video, guo2023animatediff}.
Though these methods achieve fine-grained controllability, they necessitate providing conditions for every synthetic frame, which incurs prohibitive costs in practical applications.
In this study, we aim to control video generation through temporally \emph{sparse} conditions by inputting only a few condition maps, thus making T2V more practical in a broader range of scenarios.

\para{Add-on network for additional control.}
Training foundational T2I/T2V generative models is computationally demanding.
Therefore, a preferred approach to incorporate extra control into these models is to train an additional condition encoder while maintaining the integrity of the original backbone~\cite{gal2023encoder, zhao2023uni, xu2023prompt}.
ControlNet \cite{zhang2023adding} pioneered the potential of training plug-and-play condition encoders for pre-trained T2I models.
It involves creating a trainable duplicate of the pre-trained layers that accommodates the condition input.
The encoder output is then reintegrated into the T2I model through zero-initialized layers.
Similarly, T2I-Adapter~\cite{mou2023t2i} utilizes a lightweight structure to infuse control.
IP-Adapter \cite{ye2023ip}, integrates the style condition by transposing the reference image into supplementary embeddings, which are subsequently concatenated with the text embeddings.
Our approach aligns with the principles of these works and aims to achieve sparse control through an auxiliary encoder module.

\section{SparseCtrl}
% \anyi{As we do not have a mention our method name until Sec.3.3}

\begin{figure*}
    \includegraphics[width=0.9\textwidth]{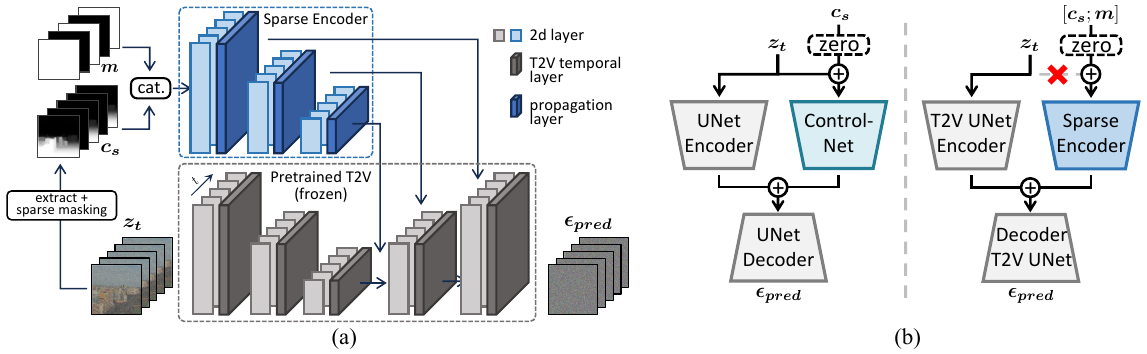}
    \vspace{5pt}
    \caption{
       (a) Overview of the \method pipeline.
       (b) Comparison between vanilla ControlNet (left) and our sparse condition encoder (right), where ``zero" stands for zero-initialized layers;
       $[\cdot; \cdot]$ denotes channel-wise concatenation.
       Detailed structures are omitted for clarity.
    }
    \vspace{-10pt}
    \label{fig:pipeline}
\end{figure*}

To enhance the controllability of a pre-trained text-to-video (T2V) model with temporally sparse signals, we introduce add-on sparse encoders to control the video generation process, leaving the original T2V generator untouched.
This section is thus organized as follows: \cref{sec:t2v} presents the background of T2V diffusion models;
\cref{sec:encoder} discuss the design of our sparse condition encoder, followed by the supported modalities and applications in \cref{sec:app}.

\subsection{Text-to-Video Diffusion Models}\label{sec:t2v}

\para{Leveraging powerful text-to-image generators.}
Text-to-image (T2I) generation has been dramatically advanced by powerful image generators like Stable Diffusion~\cite{rombach2022high}.
A practical path for T2V tasks is to 
leverage such powerful T2I priors.
Recent T2V model~\cite{blattmann2023align, ge2023preserve, zhang2023show} typically extend a pre-trained T2I generator for videos by incorporating temporal layers between the 2D image layers, as illustrated in the lower part of~\cref{fig:pipeline}~(a).
This arrangement enables cross-frame information exchange, thereby effectively modeling the cross-frame motion and temporal consistency.

\para{Training objectives.}
The training objectives of T2V models are generally aligned with their image counterparts.
Specifically, the model tries to predict the noise scale added to the clean RGB video (or latent features) $\bm{z_0^{1:N}}$ with $N$ frames, encouraged by an MSE loss:
\begin{equation}\label{eq:objective}
    \mathbb{E}_{\bm{z_0^{1:N}}, \bm{c_t}, \bm{\epsilon}, t}\left\lbrack 
\lVert \epsilon - \epsilon_\theta(\alpha_t \bm{z_0^{1:N}} + \sigma_t \bm{\epsilon}, \bm{c_t}, t) \rVert_2^2 \right\rbrack,
\end{equation}
where $\bm{c_t}$ is the embeddings of the text description, $\bm{\epsilon}$ is the sampled Gaussian noise in the same shape of $\bm{z_0^{1:N}}$, $\alpha_t$ and $\sigma_t$ are terms that control the added noise strength, $t = 1, ..., N$ is a uniformly sampled diffusion step, $T$ is the number of total steps.
In the following context, we also adopt this objective for training.

\subsection{Sparse Condition Encoder}\label{sec:encoder}
To enable efficient sparse control, we introduce an add-on encoder capable of accepting sparse condition maps as inputs, which we call sparse condition encoders.
In the T2I domain, ControlNet~\cite{zhang2023adding} successfully adds structure control to the pre-trained image generator by partially replicating a copy of the pre-train model and its input, then adding the conditions and reintegrating the output back to the original model through zero-initialized layers, as shown in the left of~\cref{fig:pipeline}~(b).
Inspired by its success, we start with a similar design to enable sparse control in the T2V setting.

\para{Limited controllability of frame-wise encoder.}\label{sec:method_framewise_enc}
We start with a straightforward solution: training a ControlNet-like encoder to incorporate sparse condition signals.
To this end, we build a frame-wise encoder akin to ControlNet, replicate it across the temporal dimension, and add the conditions to the desired keyframes through this auxiliary structure.
For frames that are not directly conditioned, we input a zero image to the encoder and indicate the unconditioned state through an additional mask channel.
However, experimental results in~\cref{sec:ablation_design} show that such frame-wise conditions sometimes fail to maintain temporal consistency when used with sparse input conditions, \emph{e.g.}, in the image animation scenario where only the first frame is conditioned.
In such cases, only keyframes react to the condition, leading to abrupt content changes between the conditioned and unconditioned frames.

\para{Condition propagation across frames.}\label{sec:method_tempora_layer}
Considering the sparsity and temporal relationship of given inputs, we hypothesize that the above problem arises because the T2V backbone has difficulty inferring the intermediate condition states for the unconditioned frames.
To solve this, we propose to add temporal layers (\emph{e.g.}, temporal attention~\cite{vaswani2017attention} with position encoding) to the sparse condition encoders that allow the conditional signal to propagate from frame to frame.
Intuitively, although not identical, different frames within a video clip share similarities in both appearance and structure.
The temporal layers can thus propagate such implicit information from the conditioned keyframes to the unconditioned frames, thereby enhancing consistency.
Our experiments confirm that this design significantly improves the robustness and consistency of the generated results.

\para{Quality degradation caused by manually noised latents.}\label{sec:method_quality_de}
Although the sparse condition encoder with temporal layers could tackle the sparsity of inputs, it sometimes leads to visual quality degradation of the generated videos, as shown in~\cref{sec:ablation_design}.
When examining the design of the vanilla ControlNet, we find that simply applying the ControlNet in our scenario is unsuitable due to the copying of noised sample inputs.
Concretely, as illustrated in~\cref{fig:pipeline}~(b), the original ControlNet copies not only the UNet~\cite{ronneberger2015unet} encoder but also the noised sample input $\bm{z_t}$.
Namely, the input for the ControlNet encoder is the sum between the condition (after zero-initialized layers) and the noised sample.
This design stabilizes the training and accelerates the model convergence in its original scenario. 
However, in terms of the unconditioned frames in our setting, the informative input of the sparse encoder becomes only the noised sample.
This might encourage the sparse encoder to overlook the condition maps and rely on the noised sample $\bm{z_t}$ during training, which contradicts our goal of controllability enhancement.
Accordingly, as shown in~\cref{fig:pipeline}~(b), our proposed sparse encoder eliminates the noised sample input and only accepts the condition maps $[\bm{c_s}, \bm{m}]$ after concatenation.
This straightforward yet effective method eliminates the observed quality degradation in our experiments.

\para{Unifying sparsity via masking.}
In practice, to unify different sparsity with a single model, we use zero images as the input placeholder for unconditioned frames and concatenate a binary mask sequence to the input conditions, which is a common practice in video reconstruction and prediction~\cite{yu2023magvit, tong2022videomae, blattmann2023align, zhang2023show, chen2023seine}.
As shown in \cref{fig:pipeline}~(a), we concatenate a mask $\bm{m} \in \{0, 1\}^{h \times w}$ channel-wise in addition to the condition signals $\bm{c_s}$ at each frame to form the input of the sparse encoder.
Setting $\bm{m}=\bm{0}$ indicates the current frame is unconditioned and vice versa.
In this way, different sparse input cases can be represented with a unified input format.

\subsection{Multiple Modalities and Applications}\label{sec:app}

In this paper, we implement \method with three modalities: sketches, depth maps, and RGB images.
Notably, our method is potentially compatible with other modalities, such as skeleton and edge map, which we leave for future developments.

\para{Sketch-to-video generation.}
Sketches~\cite{vinker2023clipascene, voynov2023sketch} can serve as an efficient guiding tool for T2V due to their ease of creation by non-professional users.
With \method, users can supply any number of sketches to shape the video content.
For instance, a single sketch can establish the overall layout of the video, while sketches of the first, last, and selected intermediate frames can define coarse motion, making the method highly beneficial for storyboarding.

\para{Depth guided generation.}
Integrating depth conditions with the pre-trained T2V enables depth-guided generation.
Consequently, users can render a video by directly exporting sparse depth maps from engines or 3D representations~\cite{mildenhall2021nerf} or conduct video translation using depth as an intermediate representation.

\para{Image animation and transition; video prediction and interpolation.}
Within the context of RGB video, numerous tasks can be unified into a single problem of video generation with RGB image conditions.
In this scheme, image animation corresponds to video generation conditioned on the first frame;
Transition is conditioned by the first and last frames;
Video prediction is conditioned on a small number of beginning frames;
Interpolation is conditioned on uniformly sparsed keyframes.

\section{Experiments}
\label{sec:exp}
In this section, we evaluate \method under various settings. 
\cref{sec:impl_details} present the detailed implementations.
\cref{sec:main_results} showcases the results and applications given one or few conditions.
\cref{sec:comparison} suggests that \method could achieve comparable performances on chosen popular tasks with baseline methods, \emph{e.g.}, sparse depth-to-video generation and image animation.
\cref{sec:ablation} present comprehensive ablation studies and evaluate \method's response to textual prompts and unrelated conditions.

\subsection{Implementation Details}\label{sec:impl_details}
\para{Text-to-video generator.}
We implement \method upon AnimateDiff~\cite{guo2023animatediff}, which can serve as a general T2V generator when integrated with its pretraining image backbone, Stable Diffusion V1.5~\cite{sd15}, or function as a personalized generator when combined with personalized image backbones such as RealisticVision~\cite{realisticvision} and ToonYou~\cite{toonyou}.
We test with both settings and showcase the results.

\para{Training.}
The training objective of \method aligns with~\cref{eq:objective}.
The only difference is the integration of the proposed sparse condition encoder into the pre-trained text-to-video (T2V) backbone.
To help the condition encoder learn robust controllability, we adopted a simple strategy to mask out conditions during training.
In each iteration, we first randomly sample a number $N_c$ between $1$ and $N$ to determine how many frames will receive the condition.
Subsequently, we draw $N_c$ indices without repeating from $\{1, 2, ..., N\}$ and keep the conditions for the corresponding frames.
We train \method on WebVid-10M~\cite{bain2021frozen} and extract the corresponding conditions on the fly.
More details can be found in the supplementary material.

\subsection{Main Results}\label{sec:main_results}

\begin{figure*}
    \includegraphics[width=0.9\textwidth]{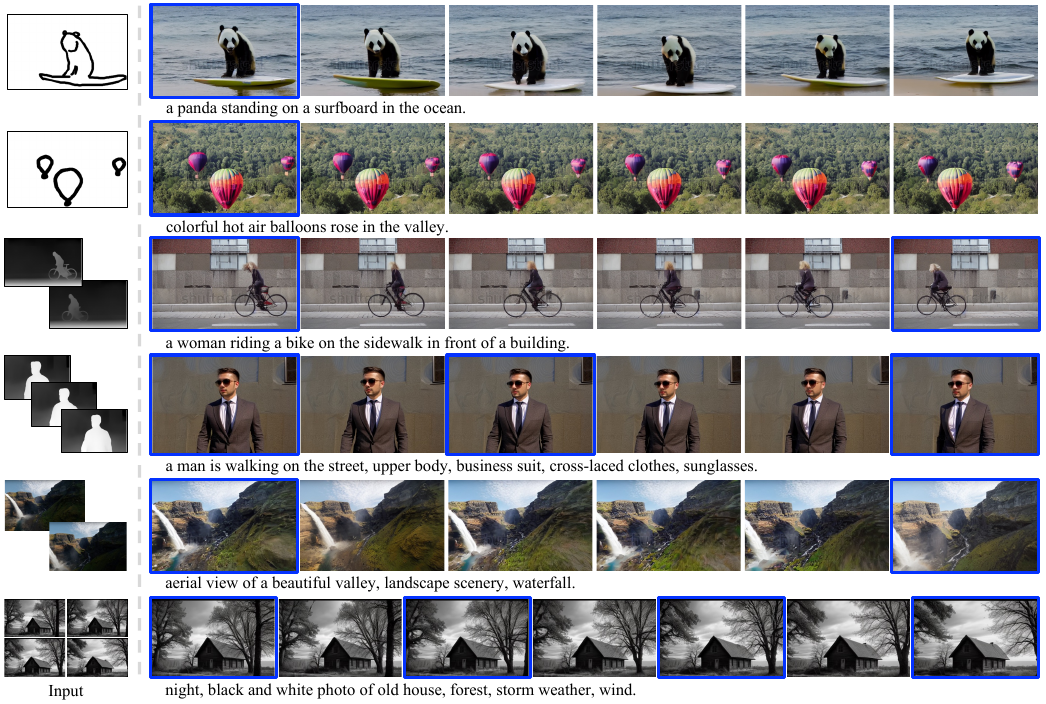}
    \caption{
        Qualitative results with sketch/depth/RGB image sparse condition encoders.
        Videos in 4/6-th rows are generated with personalized backbone, RealisticVision~\cite{realisticvision}.
        The input conditions are shown on the left; the conditioned keyframes are denoted by \textcolor{blue}{\textbf{\textit{blue}}} border.
    }
    \label{fig:qualitative}
\end{figure*}

We showcase the qualitative results and applications of \method with three modalities in~\cref{fig:teaser},~\ref{fig:qualitative}, and the supplementary material, covering original and personalized T2V settings.
As shown in the figure, with \method, the synthetic videos closely adhere to control signals and maintain an excellent temporal consistency, being robust to different numbers of conditioning frames.

Remarkably, by drawing a single sketch, we can trigger the capability of the pre-trained T2V model to generate rare semantic compositions, such as a panda standing on a surfboard shown in the first row of~\cref{fig:qualitative}.
In contrast, the pre-trained T2V model struggles to generate such complex samples using textual descriptions alone.
This suggests that the full potential of the T2V, pre-trained on large-scale datasets, may not be fully unlocked with only textual guidance.
Additionally, we show that with well-learned real-world motion knowledge, the pre-trained T2V is capable of inferring the intermediate states with as few as two conditions, as illustrated in 3/5-th rows in~\cref{fig:qualitative}.
This indicates that temporally dense control might not be necessary.

\subsection{Comparisons on Popular Tasks}\label{sec:comparison}

Since it is challenging to compare \method against prior efforts on all applications that we could enable, we choose two popular tasks for evaluation: sparse depth-to-video generation and image animation. 
For the first task, dense depth condition mode of VideoComposer (VC)~\cite{wang2023videocomposer} and Text2Video-Zero (Zero)~\cite{khachatryan2023text2video} serve as the baseline.
We also implement a baseline by combining AnimateDiff (AD)~\cite{guo2023animatediff} with ControlNet~\cite{zhang2023adding} via applying frame-wise control signals to the conditioned keyframes.
For image animation, we compare \method against two open-sourced image animation baselines: DynamiCrafter (DC)~\cite{xing2023dynamicrafter} and VideoComposer's initial frame mode.

\subsubsection{Sparse Depth-to-Video Generation}\label{sec:depth2vid}

Providing a dense depth sequence for video generation helps specify structural information to some extent. 
We thus evaluate our method on this task with much more challenging yet practical settings: only a few depths are given for the synthesis.
The controlling fidelity under different sparsity of input is measured for the quantitative comparison.
Specifically, we first select 20 videos from the validation set of WebVid-10M~\cite{bain2021frozen} that are not seen during training.
Thereafter, we estimate the corresponding depth sequences with the off-the-shelf MiDaS~\cite{ranftl2020towards} model, evenly mask out some of them with a ratio $r_{mask}$, and use the remaining depth maps as conditions to generate videos.
We then estimate the depth maps from the conditioned keyframes in generated videos and, following the metrics in previous work, we perform scale shift realignment and compute the mean absolute error (MAE) against the depth maps extracted from the original videos.
On the other hand, to prevent the model from learning a shot cut by solely controlling the keyframes and ignoring temporal consistency, we also report cross-frame CLIP~\cite{radford2021learning} similarity following previous works~\cite{wu2023tune, khachatryan2023text2video}.

\begin{table}[t]
    \centering
    \vspace{-5px}
    \resizebox{\columnwidth}{!}{
    \begin{tabular}{c cc cc cc cc}
        \toprule
        $r_{mask}$ & \multicolumn{2}{c}{$0$} & \multicolumn{2}{c}{$1/2$} & \multicolumn{2}{c}{$3/4$} & \multicolumn{2}{c}{$7/8$} \\
         & err. ($\downarrow$) & cons. ($\uparrow$) & err. & cons. & err. & cons. & err. & cons. \\
        \midrule
        VC~\cite{wang2023videocomposer}       & 8.26 & 96.02 & -    & - & - & - & - & - \\
        Zero~\cite{khachatryan2023text2video} & 8.24 & 97.05 & -    & - & - & - & - & - \\
        AD~\cite{guo2023animatediff, zhang2023adding} & 8.37 & 96.82 & 9.25 & 96.68 & 12.38 & 93.35 & 14.84 & 94.66 \\
        \textbf{Ours}                         & 8.92 & 96.54 & 8.09 & 96.75 & 7.30  & 96.48 & 7.40  & 95.56 \\
        \bottomrule
    \end{tabular}
    }
    \caption{
        Evaluation on sparse control fidelity.
        ``err." stands for MAE error; ``cons." stands for temporal consistency.
        All numbers are scaled up $100 \times$.
    }
    \label{tab:sparse}
\end{table}

\begin{figure*}[t]
    \includegraphics[width=0.9\textwidth]{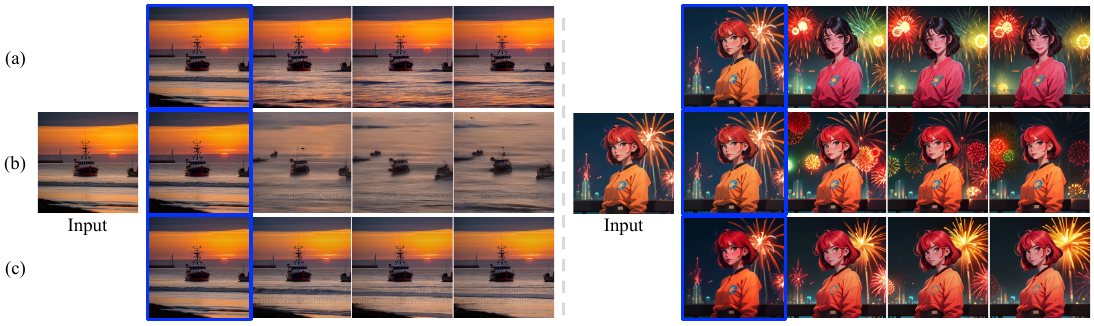}
    % \vspace{-19pt}
    \caption{
    % \anyi{TBD}
        Ablation study on network design.
        \textit{Left}: the results of wild image animation with pre-trained T2V;
        \textit{Right}: the results of \textit{in-domain} image animation with personalized T2I backbone ToonYou~\cite{toonyou}, where the input image is generated by the corresponding image model.
        The input conditions are shown on the left; the conditioned keyframes are denoted by \textcolor{blue}{\textbf{\textit{blue}}} border.
    }
    \label{fig:ablation}
\end{figure*}

The quantitative results are shown in~\cref{tab:sparse}.
To stay close to the original implementation, we only report results of $r_{mask}=0$ for VideoComposer and Text2Video-Zero, where the controls for every frame are provided.
As shown in the table, as the control sparsity, \emph{i.e.}, the masking rate $r_{mask}$, increases, our method maintains a comparable error rate with dense control baselines.
In contrast, the error of AnimateDiff with frame-wise ControlNet increases, indicating that this baseline method tends to ignore the condition signals when the control becomes sparser.

\subsubsection{Image animation}\label{sec:img2vid}

By providing the RGB image as the first frame condition, \method can handle the task of image animation.
To validate our method's effectiveness, we further compare it with two baselines in this domain.
We collect eight in-the-wild images and animate them using the three methods to generate 24 samples in total.
Similar in~\cref{sec:depth2vid}, our metrics lie in two aspects: the first frame fidelity to the input image measured by LPIPS~\cite{zhang2018unreasonable}, and temporal consistency measured by CLIP similarity.
Additionally, we invited 20 users to rank the results individually in terms of the fidelity to the given image and the overall quality preference.
We obtained 160 ranking results for each aspect.
We use average human ranking (AHR) as a preference metric and report the results in~\cref{tab:img_ani}.
The result shows that our method can achieve comparable performance with specifically designed animation pipelines while being favored in terms of fidelity to the first frame.

\begin{table}[t]
    \centering
    \vspace{-5px}
    \resizebox{\linewidth}{!}{
    \begin{tabular}{ccccc}
        \toprule
         & LPIPS ($\downarrow$) & CLIP ($\uparrow$) & fidelity(user) ($\uparrow$) & preference(user) ($\uparrow$) \\
        \midrule
        DC~\cite{xing2023dynamicrafter} & 0.5346 & 98.49 & 2.137 & 2.310 \\
        VC~\cite{wang2023videocomposer} & 0.3346 & 91.90 & 1.815 & 1.696 \\
        \textbf{Ours}                   & 0.1467 & 95.25 & 2.048 & 1.994 \\
        \bottomrule
    \end{tabular}
    }
    \caption{
        Evaluation of image animation.
    }
    \vspace{-5pt}
    \label{tab:img_ani}
\end{table}

\subsection{Ablative Study}\label{sec:ablation}

\subsubsection{Design of Sparse Encoder}\label{sec:ablation_design}

We ablate on the sparse encoder architecture to verify our choice.
Specifically, we experiment with four designs:
\textbf{(1) frame-wise condition encoder}, where we repeat the 2D ControlNet~\cite{zhang2023adding} along the temporal axis and encode the control signals to the keyframes, as depicted in~\cref{sec:method_framewise_enc};
\textbf{(2) condition encoder with propagation layers}, where we add temporal layers upon (1) to propagate conditions across frames, as discussed in~\cref{sec:method_tempora_layer};
\textbf{(3) our full model}, where we further eliminate the noised sample input to the condition encoder in (2).
To better compare the effectiveness of these three choices, we consider the most challenging case, \emph{i.e.}, the RGB image conditions, because compared to other abstract modalities, here the synthetic results need to faithfully demonstrate the fine-grained details of the condition signal and propagate it to other unconditioned frames to ensure temporal consistency.
With AnimateDiff~\cite{guo2023animatediff}, we additionally show the result on personalized image backbone, which further assists us in distinguishing the merits and shortcomings of different choices.

In~\cref{fig:ablation}, we show the qualitative image animation results.
According to the figure, with all three variations, the first frame in the generated videos is fidelity to the input image control.
The frame-wise encoder, under the personalized generation setting, fails to propagate the control to the unconditioned frames (1st row, right), leading to temporal inconsistency where the details of the character (\emph{e.g.}, hair, and clothes color) change over time.
Upon the pre-trained T2V, the encoder with propagation layers, as stated in~\cref{sec:method_quality_de}, suffers quality degradation (2nd row, left), and we hypothesize that this is because the noised sample input to the encoder provides misleading information for the condition tasks.
Finally, with propagation layers and eliminating the noised sample input, our full model works well under the two settings (3rd row), maintaining both fidelity to condition and temporal consistency.

\subsubsection{Unrelated Conditions}

\begin{figure*}[t]
    \includegraphics[width=0.9\textwidth]{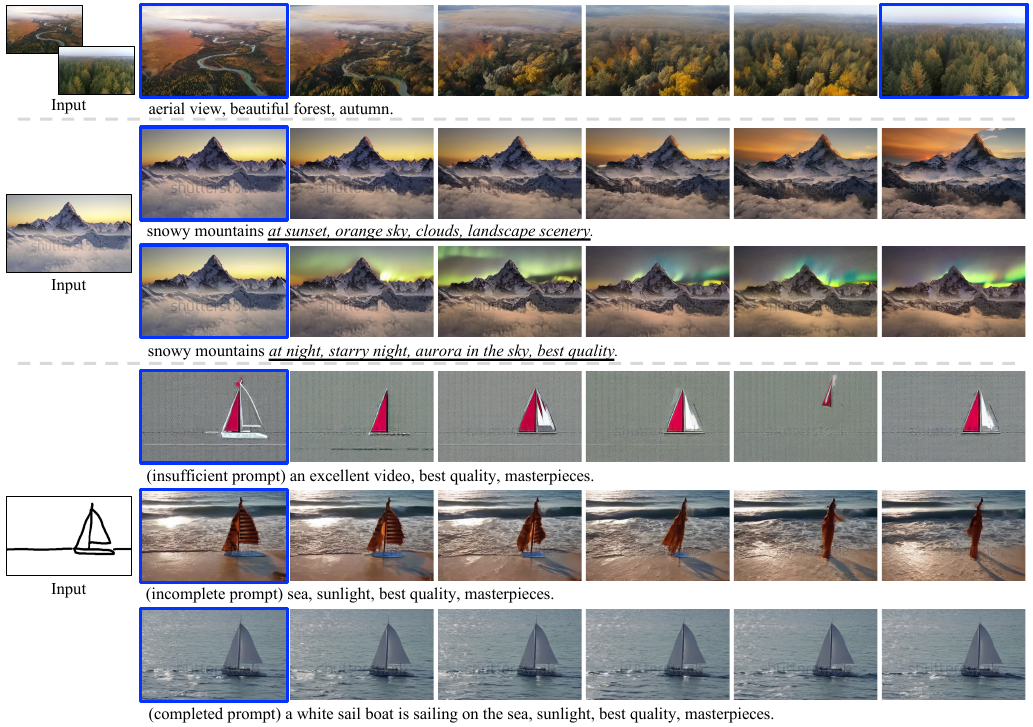}
    \vspace{-8pt}
    \caption{
    % \anyi{TBD}
        Ablation study on unrelated conditions and response to textual prompt.
        The first row demonstrates how the model deals with unrelated conditions;
        The lower five rows show how the model reacts to different textual prompts.
        The input conditions are shown on the left; the conditioned keyframes are denoted by \textcolor{blue}{\textbf{\textit{blue}}} border.
    }
    \label{fig:ablation_2}
\end{figure*}

Besides the common usages, we experiment with an extreme case where the input conditions are unrelated or contradicted.
Regarding this, we input two unrelated images to the RGB image encoder and require the model to interpolate between them, as shown in the first-row in~\cref{fig:ablation_2}.
Surprisingly, the sparse encoder can still help generate smooth transitions between the input images, which further verifies the robustness of the \method and shows potential in visual effects synthesis.

\subsubsection{Response to Textual Prompt}
Another interesting question is, with the additional information provided by the sparse condition encoder, to what extent does the final generated outcome respond to the input text description?
To answer this, we experiment with different textual prompts with the same input and demonstrate the results in~\cref{fig:ablation_2}.
In the image animation setting, we compare the prompt that faithfully describes the image content (2nd row) and the prompt that describes a slightly different content (3rd row).
The results show that the input text prompts do influence the outcome by leading the contents towards the corresponding directions.

In the sketch-to-video setting, we construct three types of prompts:
(1) insufficient prompt with no useful information (4th row), \emph{e.g.}, ``\textit{an excellent video, best quality, masterpieces}";
(2) incomplete prompt that partially describes the desired content (5th row), \emph{e.g.}, ``\textit{sea, sunlight, ...}", ignoring the central object ``sailboat";
(3) completed prompt that describes every content (6th row).
As shown in~\cref{fig:ablation_2}, with the sketch condition, the content can be properly generated only when the prompt is completed, showing that the text input still plays a significant role when the provided condition is highly abstract and insufficient to infer the content.

\section{Discussion and Conclusion}

We present \method, a unified approach of adding temporally sparse controls to pre-trained text-to-video generators via an add-on encoder network.
It can accommodate various modalities, including depth, sketches, and RGB images, greatly enhancing practical control for video generation.
This flexibility proves invaluable in diverse applications like sketch-to-video, image animation, keyframe interpolation, etc.
Extensive experiments have validated \ method's effectiveness and generalizability across original and personalized text-to-video generators, making it a promising tool for real-world usage.

\para{Limitations.}
Though with \method, the visual quality, semantic composition ability, and domain of the generated results are limited by the pre-trained T2V backbone and the training data.
In experiments, we find that the failure cases mostly come from out-of-domain input, such as anime image animation, since such data is scarce in the T2V and sparse encoder's pre-training dataset WebVid-10M~\cite{bain2021frozen}, whose contents are mainly real-world videos.
Possible solutions for enhancing the generalizability could be improving the training dataset's domain diversity and utilizing some domain-specific backbone, such as integrating \method with AnimateDiff~\cite{guo2023animatediff}.

\para{Acknowledgement.}
The project is supported by the Shanghai Artificial Intelligence Laboratory (P23KN00601, P23KS00020, 2022ZD0160201), CUHK Interdisciplinary AI Research Institute, and the Centre for Perceptual and Interactive Intelligence (CPIl) Ltd under the Innovation and Technology Commission (ITC)'s InnoHK.

{
    \small
    \bibliographystyle{ieeenat_fullname}
    \bibliography{main}
}

\end{document}